\pdfoutput=1
\documentclass[11pt]{article}
\usepackage{ACL2023}

\usepackage{times}
\usepackage{latexsym}
\usepackage[T1]{fontenc}
\usepackage{soul}
\usepackage[utf8]{inputenc}
\usepackage{microtype}
\usepackage{inconsolata}
\usepackage{diagbox}
\usepackage{graphicx}
\usepackage{makecell}
\usepackage{multirow}
\usepackage{booktabs}
\usepackage{listings}
\usepackage{svg}
\usepackage{pdfpages}

\title{Automated Refugee Case Analysis:\\An NLP Pipeline for Supporting Legal Practitioners}

\author{Claire Barale \and Michael Rovatsos \\ School of Informatics \\ The University of Edinburgh \\ {\texttt{\{claire.barale,michael.rovatsos\}@ed.ac.uk}} \And Nehal Bhuta \\ School of Law \\ The University of Edinburgh \\ \texttt{nehal.bhuta@ed.ac.uk}}

\begin{document}
\maketitle

\begin{abstract}
In this paper, we introduce an end-to-end pipeline for retrieving, processing, and extracting targeted information from legal cases. We investigate an under-studied legal domain with a case study on refugee law in Canada. 

Searching case law for past similar cases is a key part of legal work for both lawyers and judges, the potential end-users of our prototype. While traditional named-entity recognition labels such as dates provide meaningful information in legal work, we propose to extend existing models and retrieve a total of 19 useful categories of items from refugee cases.

After creating a novel data set of cases, we perform information extraction based on state-of-the-art neural named-entity recognition (NER). We test different architectures including two transformer models, using contextual and non-contextual embeddings, and compare general purpose versus domain-specific pre-training. 

The results demonstrate that models pre-trained on legal data perform best despite their smaller size, suggesting that domain matching had a larger effect than network architecture. We achieve a F1 score above 90\% on five of the targeted categories and over 80\% on four further categories.

\end{abstract}

\section{Introduction} \label{sec: intro}
The retrieval of similar cases and their analysis is a task at the core of legal work. Legal search tools are widely used by lawyers and counsels to write applications and by judges to inform their decision-making process. However, this task poses a series of challenges to legal professionals: (i) it is an expensive and time-consuming task that accounts for 30\% of the legal work on average \cite{poje2014legal}, (ii) databases can be very large, with legal search tools gathering billions of documents, and (iii) selection of cases can be imprecise and may return many irrelevant cases, which creates the need to read more cases than necessary.

In Canada, from the date of the first claim to the final decision outcome, a claimant can expect to wait 24 months for refugee claims and 12 months for refugee appeals\footnote{Wait times for refugee claims in Canada: \url{https://irb.gc.ca/en/transparency/pac-binder-nov-2020/Pages/pac8a.aspx?=undefined&wbdisable=true}}. Long processing times are due to a significant backlog and to the amount of work required from counsels that help claimants file their claims, and who are frequently legal aid or NGO employees.

We find that these challenges are well-suited for NLP-based solutions and investigate the feasibility of automating all steps of the legal search for past similar cases. We construct an end-to-end pipeline that aims at facilitating this multi-step process, thereby supporting and speeding up the work of both lawyers and judges in \textit{Refugee Status Determination (RSD)}. We provide a level of granularity and precision that goes beyond that of existing legal search tools such as \textit{Westlaw}, \textit{LexisNexis}, or \textit{Refworld}\footnote{Refworld: \url{https://www.refworld.org/}} \cite{custis2019westlaw}, which operate at the document level. \textit{Refworld} is an online database maintained by the United Nations which helps retrieve relevant precedent cases and legislation. However, the level of precision with which one can search for cases is limited. Moreover, our pipeline guarantees increased transparency, enabling end users to choose the criteria of legal search they find most relevant to their task among the proposed categories that act as filters for a search.

\begin{figure*}
   \includegraphics[width=\textwidth]{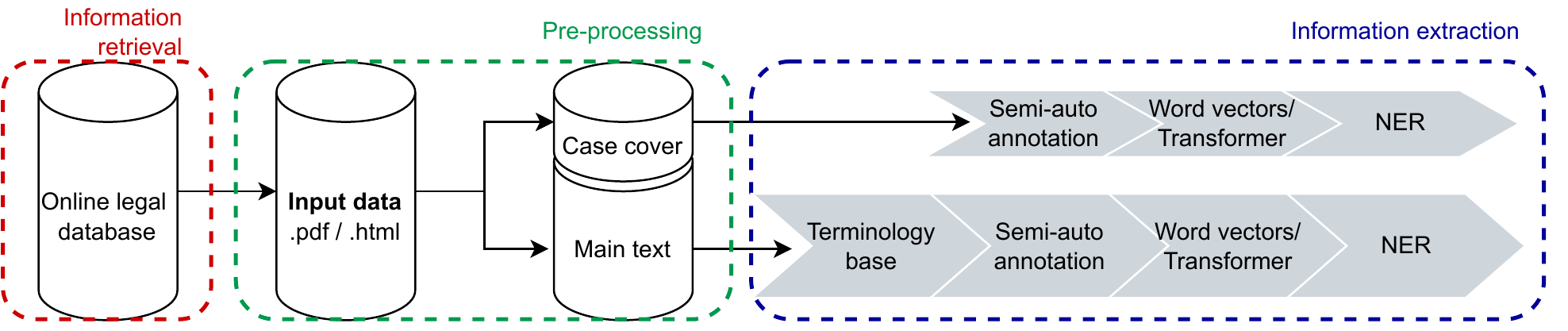} 
   \centering
   \caption{End-to-end automated pipeline}
   \label{figure: pipeline}
\end{figure*}

Specific literature studying refugee law and AI is sparse. Attention has been given to the classification and prediction of asylum cases in the United States \cite{chen_asylum_2017, dunn_early_2017}. On Canadian data, research has been conducted to analyze the disparities in refugee decisions using statistical analysis \cite{rehaag2007troubling, rehaag2019judicial, cameron_artificial_2021}. However, those studies rely mostly on tabular data. We propose to work directly on the text of refugee cases. To the best of our knowledge, no previous work implements an end-to-end pipeline and state-of-the-art NLP methods in the field of refugee law. 

We provide an NLP-based end-to-end prototype for automating refugee case analysis built on historical (already decided) cases, which are currently available only in unstructured or semi-structured formats, and which represent the input data to our pipeline. The end goal of our approach is to add structure to the database of cases by extracting targeted information described in table \ref{table: target items} from the case documents, and providing the results in a structured format to significantly enrich the search options for cases. Thereby, the input data set of cases is described in a structured manner based on our extracted categories of items, adding extensive capabilities for legal search. 

The pipeline described in figure \ref{figure: pipeline} begins by searching and downloading cases (information retrieval, paragraph \ref{subsec:4.1 IR}), pre-processing them (paragraph \ref{subsection:4.2 pre-processing}), extracting items previously identified as relevant by legal professionals. It then outputs a structured, precise database of refugee cases (information extraction, paragraph \ref{subsection:4.3 IE}). In the information extraction step, we test different training and pre-training architectures in order to determine the best methods to apply to the refugee case documents. We construct each step with the aim of minimizing the need for human effort in creating labeled training data, aiming to achieve the best possible accuracy on each extracted information item. We discuss technical choices and methodologies in section \ref{sec: 5-methodology}. Finally, we evaluate the information extraction step on precision, recall, and F1 score, and present detailed results in section \ref{sec: 6- results}.


We demonstrate that annotation can be sped up by the use of a terminology base while incorporating domain knowledge and semi-automated annotation tools. We find that domain matching is important for training to achieve the highest possible scores. We reach satisfactory token classification results on a majority of our chosen categories. The contributions of this paper are as follows:
\begin{enumerate}
\item First, we retrieve 59,112 historic decision documents (dated from 1996 to 2022) from online services of the Canadian Legal Information Institute (CanLII) based on context-based indexing and metadata to curate a collection of federal \textit{Refugee Status Determination} (RSD) cases. Our automated retrieval process is exhaustive and comprises all available cases. It is superior to human-based manual retrieval in terms of error proneness and processing time.
\item Second, we proposed an information extraction pipeline that involves pre-processing, construction of a terminology base, labeling data, and using word vectors and NER models to augment the data with structured information. We fine-tune state-of-the-art neural network models to the corpus of our retrieved cases by training on newly created gold-standard text annotations specific to our defined categories of interest.
\item Lastly, we extract the targeted category items from the retrieved cases and create a structured database from our results. We introduce structure to the world of unstructured legal \textit{RSD} cases and thereby increase the transparency of stated legal grounds, judge reasoning, and decision outcomes across all processed cases. 

\end{enumerate}

\begin{table*}[h!]
\centering
\resizebox{\textwidth}{!}{
\begin{tabular}{|c|c|c|c|c|}
\hline
 & \textbf{Label} & \textbf{A} & \textbf{Description} & \textbf{Example} \\

\hline

 \multirow{5}{*}{\rotatebox[origin=c]{90}{\textbf{Case cover}}} & \multicolumn{4}{c|}{\textbf{\textit{General}}} \\
 \cline{2-5}
 & \texttt{DATE} & 1,219 & absolute or relative dates or periods & date of the hearing and date of the decision \\
 & \texttt{GPE} & 871 & cities, countries, regions & place of the hearing\\
 & \texttt{ORG} & 278 & tribunals & "immigration appeal division", "refugee protection division" \\
 & \texttt{PERSON} & 119 & names & name of the panel and counsels\\
 \hline 
 \multirow{20}{*}{\rotatebox[origin=c]{90}{\textbf{Main text}}} &\multicolumn{4}{c|}{\textbf{\textit{Information on claimant and allegations}}} \\
 \cline{2-5}
 & \texttt{CLAIMANT\_EVENT} & 1,575 & verbs or nouns describing an event of the story of the claimant & "rape", "threat", "attacks", "fled"\\
 & \texttt{CLAIMANT\_INFO} & 235 & age, gender, citizenship, occupation & "28 year old", "citizen of Iran", "female" \\
 & \texttt{GPE} & 732 & cities, countries, regions & countries of past residency or places of hearings: "toronto, ontario" \\
 & \texttt{NORP} & 129 & nationalities, religious, political or ethnic groups or communities & "hutu", "nigerian", "christian" \\
 \cline{2-5}
 & \multicolumn{4}{c|}{\textbf{\textit{Legal procedure}}} \\
 \cline{2-5}
 & \texttt{ORG} & 549 & tribunals, NGOs, companies & "human rights watch", "refugee protection division" \\
 & \texttt{PROCEDURE} & 594 & steps in the claim and legal procedure events & "removal order", "sponsorship for application" \\
 \cline{2-5}
 & \multicolumn{4}{c|}{\textbf{\textit{Analysis and reasons for decision outcome}}} \\
 \cline{2-5}
 & \texttt{CREDIBILITY} & 684 & mentions of credibility in the determination & "lack of evidence", "inconsistencies" \\
 & \texttt{DETERMINATION} & 76 & outcome of the decision (accept/reject) & "appeal is dismissed", "panel determines that the claimant is not a convention refugee" \\
 & \texttt{DOC\_EVIDENCE} & 768 & pieces of evidence, proofs, supporting documents & "passport", "medical record", "marriage certificate" \\
 & \texttt{EXPLANATION} & 404 & reasons given by the panel for the determination & "fear of persecution", "no protection by the state" \\
 \cline{2-5}
 & \multicolumn{4}{c|}{\textbf{\textit{Timeline}}} \\
 \cline{2-5}
 & \texttt{DATE} & 628 & absolute or relative dates or periods  & "for two ears", "june, 4th 1996" \\
 \cline{2-5}
 & \multicolumn{4}{c|}{\textbf{\textit{Names}}}\\
 \cline{2-5}
 & \texttt{PERSON} & 154 & names & claimants' names, their family, name of judges\\
 \cline{2-5}
 & \multicolumn{4}{c|}{\textbf{\textit{Citations}}} \\
 \cline{2-5} 
 & \texttt{LAW} & 476 & legislation and international conventions & state law and international conventions, "section 1(a) of the convention" \\
 & \texttt{LAW\_CASE} & 109 & case law and past decided cases, by the same tribunal or another & "xxx v. minister of canada, 1994" \\
 & \texttt{LAW\_REPORT} & 18 & country reports written by NGOs or the United Nations & " 	amnesty international, surviving death: police and military torture of women in mexico, 2016" \\

\hline

\end{tabular}}
\caption{Targeted categories (\textbf{Label}) for extraction with number of annotations (\textbf{A}) per label, sorted alphabetically}
\label{table: target items}
\end{table*}

\section{Background and motivation}
At the core of the ongoing refugee crisis is the legal and administrative procedure of \textit{Refugee Status Determination} (\textit{RSD}), which can be summarized in three sub-procedures: (i) a formal claim for refugee protection by a claimant who is commonly supported by a lawyer, (ii) the decision-making process of a panel of judges and (iii) the final decision outcome with written justification for granting refugee protection or not. 

Refugee protection decisions are high-stakes procedures that target 4.6 million asylum seekers worldwide as of mid-2022. In Canada alone, 48,014 new claims and 10,055 appeals were filed in 2021\footnote{\url{https://irb.gc.ca/en/statistics/Pages/index.aspx}}. As stated in the introduction, processing times of refugee claims vary and range from a few months to several years. One of the reasons for the long processing times is the effort required for similar cases search. Case research is an essential part of the counsel's work in preparation for a new claim file. This search involves retrieving citations and references to previous, ideally successful RSD cases that exhibit similarities to the case in preparation such as the country of origin or the reason for the claim. Equally, judges rely on researching previous cases to justify their reasoning and ensure coherency across rulings. 

While each case exhibits individual characteristics and details, legal practitioners typically search for similarities based on the constitution of the panel, the country of origin and the characteristics of the claimant, the year the claim was made in relation to a particular geopolitical situation, the legal procedures involved, the grounds for the decision, the legislation, as well as other cases or reports that are cited. 

Our work aims to support legal practitioners, both lawyers preparing the application file and judges having to reach a decision, by automating the time-consuming search for similar legal cases referred to here as \textit{refugee case analysis}. As a case study, we work on first instance and appeal decisions made by the \textit{Immigration and Refugee Board of Canada}. A common approach used by legal practitioners is to manually search and filter past RSD cases on online services such as CanLII or Refworld by elementary \textit{document text} search, which is a keyword-based \textit{find exact} search, or by date. 

Our defined categories of interest are described in table \ref{table: target items}. The labels have been defined and decided upon with the help of three experienced refugee lawyers. From the interviews, we curated a list of keywords, grounds, and legal elements determining a decision. Moreover, we analyzed a sample of 50 Canadian refugee cases recommended by the interviewees to be representative over years of the claim and tribunals.

We use the pre-defined labels provided by \texttt{spaCy}'s state-of-the-art \texttt{EntityRecognizer} class including \texttt{DATE}, \texttt{PERSON}, \texttt{GPE}, \texttt{ORG}, \texttt{NORP}, \texttt{LAW} and extend this list with new additional labels that we created and trained from scratch. 

Each case document comprises a \textit{case cover} page (the first page) and the \textit{main text} which differ in the type and format of their information content. Therefore, we chose separate labels for the \textit{case cover}. The \textit{case cover} contains general information about the case (cf.\ example in Appendix \ref{sec:appendix A case cover}). While the main text is presented as a full-body text, the \textit{case cover} page consists of semi-structured information which could that could be roughly described as tabular, except it does not follow a clear layout. Based on the \textit{case cover} page we aim to extract meta-information about each claim using four labels (table \ref{table: target items}).

For the \textit{main text}, we chose 15 labels that represent characteristics reflective of similarity among different cases. To link cases to each other and later facilitate similar case retrieval, we also extract three categories of citations i.e. \texttt{LAW} for legal texts, \texttt{LAW\_CASES} for other mentioned past cases, and \texttt{LAW\_REPORT} for external sources of information such as country reports. Additionally, the \texttt{CREDIBILITY} label retrieves mentions made of credibility concerns in the claimant's allegations, which tends to be among the deciding factors for the success of a claim and is hence essential to understand the reasoning that led to the legal determination at hand.
 


A successful implementation of a system capable of extracting this information reliably would provide several benefits to legal practitioners: (i) facilitating, speeding up, and focusing legal search, (ii) reducing the time spent on a claim and on providing relevant references, potentially resulting in a file that has more chances of being accepted, and (iii) for judges, to ensure consistent outcomes across time and different jurisdictions or claimant populations. 

\section{Research approach} \label{sec: research approach}
Our approach is guided by investigating the hypothesis that NER methods can be used to extract structured information from legal cases, i.e.\ we want to determine whether state-of-the-art methods can be used to improve the transparency and processing of refugee cases. Consistency of the decision-making process and thorough assessment of legal procedure steps are crucial aspects ensuring that legal decision outcomes are transparent, high-quality, and well-informed. Consequently, key research questions we need to address include:

\noindent\textbf{Training data requirements}  How many labeled samples are needed? Can keyword-matching methods or terminology databases be leveraged to reduce the need for human annotation?

\noindent\textbf{Extraction} 
What methods are best suited to identify and extract the target information from legal cases? 

\noindent\textbf{Replicability}  To what extent might our methods generalize to other legal data sets (other legal fields or other jurisdictions)?

\noindent\textbf{Pre-training}  How important is the pre-training step? How important is {\em domain matching}: do domain-specific pre-training perform better than general-purpose embeddings, despite their smaller sizes?
    
\noindent\textbf{Architectures}  How important is the architecture applied to the information extraction tasks, in terms of F1 score, precision, and recall?



\section{Pipeline details and experimental setup} \label{sec: 4 pipeline details}
In this section, we detail each step of the pipeline as presented in figure \ref{figure: pipeline} and how it compares to the current legal search process. Subsequently, in \ref{sec: 5-methodology} we describe the training data creation process, and the network architectures tested. The code for our implementation and experiments can be found at \url{https://github.com/clairebarale/refugee_cases_ner}.

\subsection{Information retrieval: case search} \label{subsec:4.1 IR}
We retrieve 59,112 cases processed by the \textit{Immigration and Refugee Board} of Canada that range from 1996 to 2022. The case documents have been collected from CanLII in two formats, PDF and HTML. The CanLII web interface serves queries through their web API accessible at the endpoint with URL \url{https://www.canlii.org/en/search/ajaxSearch.do}. For meaningful queries, the web API exposes a number of HTTP-GET request parameters and corresponding values which are to be appended to the URL but preceded by a single question mark and then concatenated by a single ampersand each. For instance, in the \texttt{parameter=value} pairs in the following example, the keyword search exactly matches the text \textit{REFUGEE}, and we retrieve the second page of a paginated list of decisions from March 2004 sorted by descending date, which returns a JSON object (Full query: \url{https://www.canlii.org/en/search/ajaxSearch.do?type=decision&ccId=cisr&text=EXACT(REFUGEE)&startDate=2004-03-01&endDate=2004-03-31&sort=decisionDateDesc&page=2}). Note that CanLII applies pagination to the search results in order to limit the size of returning objects per request.


\subsection{Preprocessing} \label{subsection:4.2 pre-processing}
We obtain two sets: (1) a set of \textit{case covers} that consists of semi-structured data and displays meta-information and (2) a set of \textit{main text} that contains the body of each case, in full text.

Generally, the CanLII database renders the decision case documents as HTML pages for display in modern web browsers but also provides PDF files. We use \texttt{PyPDF2}\footnote{\texttt{PyPDF2}: \url{https://github.com/py-pdf/PyPDF2}} for parsing the contents of PDF files as text. To parse the contents of HTML files as text input to our NLP pipeline, we use the {\texttt{BeautifulSoup}\footnote{\texttt{BeautifulSoup}: \url{https://beautiful-soup-4.readthedocs.io/en/latest/}} python library.

The choice between PDF and HTML format is based on multiple reasons, as each format has its own advantages and disadvantages. First, depending on the text format \texttt{PyPDF2} occasionally adds excessive white space between letters of the same word. Also, the PDF document is parsed line-by-line from left to right, top to bottom. Therefore, multi-column text is often mistakenly concatenated as a single line of text. However, the available PDF documents are separated by pages and \texttt{PyPDF2} provides functionality to select individual document pages which we used to select the case cover page that provides case details for each document. HTML as markup language provides exact anchors with HTML tags, which, in most cases, are denoted by opening and closing tag parts such as \texttt{<p>} and \texttt{</p>} for enclosing a paragraph.

When processing the \textit{main text} of each case document, we parse the HTML files using \texttt{BeautilfulSoup}, remove the case cover to keep only the full-body text, and tokenize the text by sentence using the \texttt{NLTK}\footnote{\texttt{NLTK tokenizer}: \url{https://www.nltk.org/api/nltk.tokenize.html}}. Our preference to tokenize by sentence facilitates the annotation process while keeping the majority of the context. We also experimented with splitting by paragraph which yielded relatively large chunks of text, whereas splitting by phrase did not keep enough context during the annotation process. To gather results, we create a \texttt{pandas} \texttt{Dataframe}, create a sentence per row, and save it to a CSV file.

For the \textit{case cover}, we exploit \texttt{PyPDF2}'s functionality to extract the text of the first page from the PDF format. In contrast to this, when using \texttt{BeautifulSoup} we could not rely on HTML tags (neither through generic tag selection nor by CSS identifier (ID) or CSS class), to retrieve the first page of the document robustly. After extracting this page for each case, we parse the PDF files as plain text. Combined with the metadata from the document retrieval provided by CanLII, we derive the case identifier number and assign it to the corresponding PDF file. As a next step and similar to the procedure for the main body of each document, we create a \texttt{pandas Dataframe} from the extracted data and save it as a CSV file with case identifier numbers and their associated case cover.

For both file formats, we perform basic text cleaning, converting letters to lowercase, and removing excessive white space and random newlines.

\subsection{Information extraction} \label{subsection:4.3 IE}
The goal of our pipeline is not only to retrieve the cases but to structure them with a high level of precision and specificity, and to output a tabular file where each column stores specific  information of each of our target types for each case. Using such structured information, legal practitioners can find similar cases with ease instead of reading carefully through several cases before finding a few cases similar to their own cases by selecting attributes in one or several of the extracted categories.

We chose to use neural network approaches to perform the information extraction step. After some experimentation, approaches such as simple matching and regular expressions search proved too narrow and unsuitable for our data. Given the diversity of formulations and layouts, phrasing that captures context is quite important. Similarly, we discard unsupervised approaches based on the similarity of the text at the document or paragraph level because we favor transparency to the end user in order to enable leveraging legal practitioners' knowledge and expertise.

Extraction of target information can be done using sequence-labeling classification. NER methods are well-suited to the task of extracting keywords and short phrases from a text. To this end, we create a training set of annotated samples as explained in the next section \ref{subsec:5.1 training data}. Labeled sentences are collected in jsonlines format, which we convert to the binary \texttt{spaCy}-required format and use as training and validation data for our NER pipeline.

\section{Methodology} \label{sec: 5-methodology}

\subsection{Training data creation} \label{subsec:5.1 training data}
We choose to use a machine learning component for text similarity to reinforce the consistency of the annotations. In line with our previous step of pre-processing, we annotate the case cover section and the main text separately. While we decided to annotate the whole page of the case cover because the semi-structured nature of the text makes tokenization approximate, we perform annotation of the main text as a sentence-based task, preserving some context. We use the \texttt{Prodigy} annotation tool\footnote{\texttt{Prodigy}: \url{https://prodi.gy/docs}}, 
which provides semi-automatic annotations and active learning in order to speed up and improve the manual labeling work in terms of consistency and accuracy of annotation. We convert the two \texttt{pandas Dataframes} containing the pre-processed text to jsonlines which is the preferred format for \texttt{Prodigy}. We annotate 346 case covers and 2,436 sentences for the main text, which are chosen from the corpus at random. 

To collect annotated samples on traditional NER labels (\texttt{DATE, ORG, GPE, PERSON, NORP, LAW}), we use suggestions from general purpose pre-trained embeddings\footnote{\url{https://spacy.io/models/en}}. For the remaining labels (\texttt{CLAIMANT\_INFO, CLAIMANT\_EVENT, PROCEDURE, DOC\_EVIDENCE, EXPLANATION, DETERMINATION, CREDIBILITY}), and still with the aim of improving consistency of annotation, we create a terminology base (as shown on pipeline description figure \ref{figure: pipeline}). At annotation time, patterns are matched with shown sentences. The human annotator only corrects them, creating a gold standard set of
sentences and considerably speeding up the labeling task.

To create a terminology base for each target category, we first extract keywords describing cases from CanLII metadata retrieved during the information retrieval step. To this initial list of tokens, we add a list of tokens that were manually flagged in cases by legal professionals. We delete duplicates and some irrelevant or too general words such as ``claimant'' or ``refugee'', and manually assign the selected keywords to the appropriate label to obtain a list of tokens and short phrases per label. In order to extend our terminology base, we use the \texttt{sense2vec} model\footnote{\texttt{Sense2vec}: \\ \url{https://github.com/explosion/sense2vec}}(based on \texttt{word2vec} \cite{mikolov2013efficient}) to generate similar words and phrases. We select every word that is at least 70\% similar to the original keyword in terms of cosine similarity and obtain a JSON file that contains 1,001 collected patterns. This method allows us to create a larger number of labeled data compared to fully manual annotation in the same amount of time.

Table \ref{table: target items} describes the breakdown of labels in our annotated data. There is a clear imbalance across categories of items, with some labels being infrequent (\texttt{NORP, DETERMINATION, PERSON, LAW\_REPORT, LAW\_CASE}). Some labels are present very few times per case: \texttt{DETERMINATION} occurs only once per case, \texttt{PERSON} does not occur frequently since most cases are anonymized.


\subsection{Experimental conditions and architectures}

\noindent\textbf{Train, dev, test split} We trained the NER models using 80\% of the labeled data as our training set (276 case covers and 1,951 sentences for the main text, respectively), 10\% of the labeled data as our development set (35 case covers and 244 sentences) and 10\% of the labeled data as the test set for evaluation (35 case covers and 244 sentences).
\\

\noindent\textbf{Pre-training static and contextual embeddings}
As the first layer of the NER network, we add pre-trained character-level embeddings in order to isolate the effect of pre-training from the effect of the architecture and improve the F1 score on target items. We fine-tune \texttt{GloVe} vectors (\cite{pennington2014glove}, 6B tokens, 400K vocabulary, uncased, 50 dimensions) on our data using the \texttt{Mittens}\footnote{\texttt{Mittens}: \url{https://github.com/roamanalytics/mittens}}
python package \cite{dingwall-potts-2018-mittens} and create 970 static vectors. On top of the generated static vectors, we add dynamic contextualized vectors using pre-training embeddings based on \texttt{BERT} \cite{devlin2019bert}, updating weights on our corpus of cases. Because the text of the case cover is presented in a semi-structured format, we consider that it is unnecessary to perform pre-training  given the lack of context around the target items.
\\

\noindent\textbf{Architectures}
We experiment with five different architectures on the case cover and seven different architectures on the main text: five based on convolutional neural networks (CNN) using different word embeddings and two transformer architectures. We train a CNN without added vectors as a baseline. Only the transformer architectures require training on a GPU. We use the \texttt{spaCy} pipelines\footnote{\texttt{SpaCy}: \url{https://spacy.io/api/entityrecognizer}} (tokenizer, CNN and transformer) and the \texttt{HuggingFace} datasets\footnote{\texttt{roBERTa}: \url{https://huggingface.co/roberta-base}, \texttt{LegalBERT}:\url{https://huggingface.co/nlpaueb/legal-bert-base-uncased}}. All CNNs use an \texttt{Adam} optimizer function. Since the sentence-labeling task is well-suited to the masked language modeling objective, we chose to experiment with \texttt{roBERTa} \cite{liu2019roberta} and \texttt{LegalBERT} \cite{chalkidis-etal-2020-legal} in order to compare performance between a general content and a legal content model.

We train separately on the case cover, the traditional NER labels (\texttt{GPE, NORP, ORG, DATE, PERSON, LAW}), and the labels we created from scratch since it was observed that labels trained from scratch benefit from a lower learning rate (0.0005 versus 0.001 for the traditional labels).

\section{Results and evaluation} \label{sec: 6- results}

Our experimental results are presented in table \ref{tab:results table} in absolute terms and relative to the baseline in figure \ref{fig:baseline_graph} below. Our chosen baseline is a CNN with no additional vectors. We present them per label because of the disparities in the scores. The upper rows contain results on the case cover and the lower rows results on the main text. The evaluation metrics applied serve a duel purpose: for future research, achieving a high F1 score and precision-recall balance is key, while for our potential legal end users we assume that the recall measure is much more important as it measures how many of the predicted entities are correct.

\begin{figure}
   \includegraphics[width=\columnwidth]{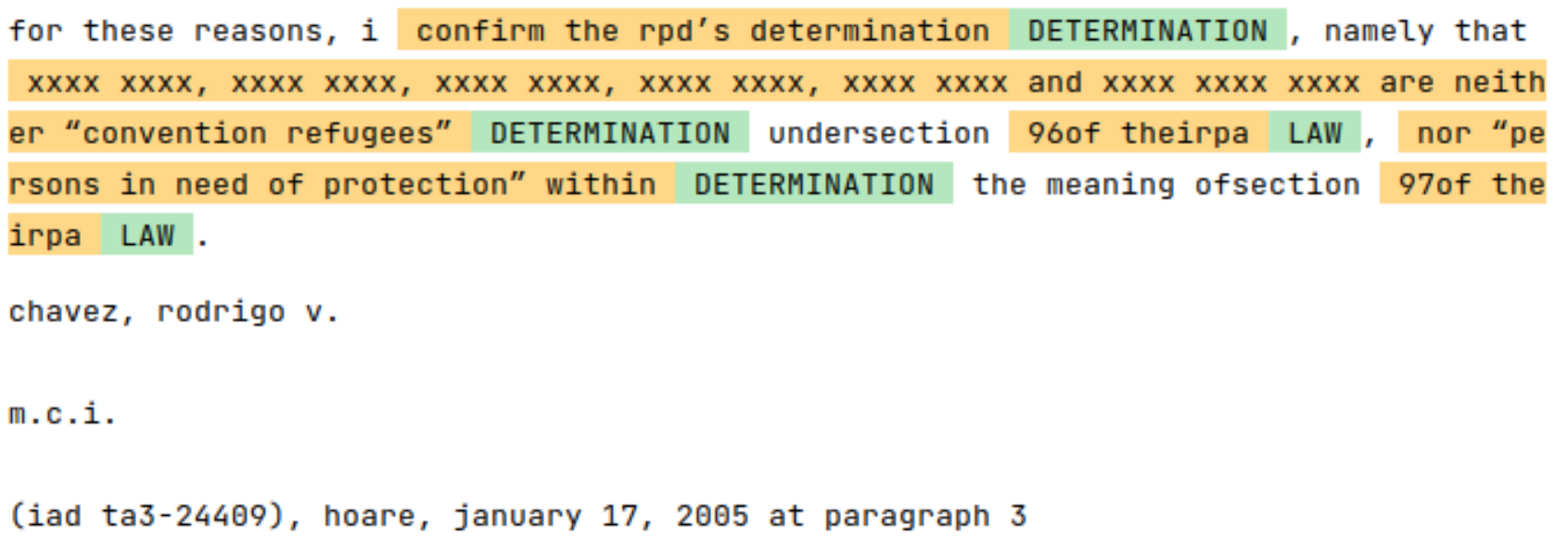} 
   \centering
   \caption{Example of an error in tokenization}
   \label{figure: tokenization}
\end{figure}

\begin{table*}[h!]
\centering
\resizebox{\textwidth}{!}{
\begin{tabular}{c|
c|c|c| 
c|c|c|
c|c|c| 
c|c|c| 
c|c|c| 
c|c|c|
c|c|c} 

\toprule

\diagbox{\textbf{Label}}{\textbf{Architecture}} 
& \multicolumn{3}{c|}{\texttt{baseline}} 
& \multicolumn{3}{c|}{\texttt{CNN+rsv}} 
& \multicolumn{3}{c|}{\texttt{CNN+fts}} 
& \multicolumn{3}{c|}{\texttt{CNN+rsv+pt}}
& \multicolumn{3}{c|}{\texttt{CNN+fts+pt}}
& \multicolumn{3}{c|}{\texttt{RoBERTa}}
& \multicolumn{3}{c}{\texttt{LegalBERT}} \\

\midrule


& \textbf{P} & \textbf{R} & \textbf{F1}
& \textbf{P} & \textbf{R} & \textbf{F1}
& \textbf{P} & \textbf{R} & \textbf{F1}
& \textbf{P} & \textbf{R} & \textbf{F1}
& \textbf{P} & \textbf{R} & \textbf{F1}
& \textbf{P} & \textbf{R} & \textbf{F1}
& \textbf{P} & \textbf{R} & \textbf{F1}\\

\midrule
\multicolumn{22}{c}{\textbf{Header and cover page}} \\
\midrule


\texttt{DATE} & 97.46 & 95.04 & 96.23 & 95.90 & 96.69 & \textbf{96.30} & - & - & - & - & - & - & - & - & - & 89.92 & 89.17 & 89.54 & 89.08 & 88.33 & 88.70\\ 
\texttt{GPE} & 92.96 & 91.67 & \textbf{92.31} & 90.14 & 88.89 & 89.51 & - & - & - & - & - & - & - & - & - & 90.54 & 93.06 & 91.78 & 88.00 & 91.67 & 89.80\\ 
\texttt{ORG} & 94.74 & 90.00 & \textbf{92.31} & 94.74 & 90.00 & \textbf{92.31} & - & - & - & - & - & - & - & - & - & 79.17 & 95.00 & 86.36 & 86.36 & 95.00 & 90.48\\ 
\texttt{PERSON} & 80.80 & 84.17 & 82.45 & 81.75 & 85.83 & 83.74 & - & - & - & - & - & - & - & - & - & 75.48 & 96.69 & \textbf{84.78} & 69.82 & 97.52 & 81.38\\ 


\midrule
\multicolumn{22}{c}{\textbf{Main text document body}} \\
\midrule
\texttt{CLAIMANT\_EVENT} & 60.36 & 44.67 & 51.34 & 57.02 & 46.00 & 50.92 & 57.89 & 36.67 & 44.90 & 55.04 & 47.33 & 50.90 & 63.71 & 52.67 & 57.66 & 64.34 & 61.33 & 62.80 & 65.10 & 64.67 & \textbf{64.88}  \\  
\texttt{CLAIMANT\_INFO}  & 55.00 & 61.11 & 57.89 & 47.83 & 61.11 & 53.66 & 61.11 & 61.11 & 61.11 & 55.56 & 55.56 & 55.56 & 63.16 & 66.67 & \textbf{64.86} & 63.16 & 66.67 & \textbf{64.86} & 57.89 & 61.11 & 59.46  \\  
\texttt{CREDIBILITY}    & 68.57 & 50.00 & 57.83 & 62.50 & 52.08 & 56.82 & 69.23 & 56.25 & \textbf{62.07} & 74.19 & 47.92 & 58.23 & 68.42 & 54.17 & 60.47 & 56.60 & 62.50 & 59.41 & 62.50 & 52.08 & 56.82  \\  
\texttt{DATE}           & 72.34 & 69.39 & 70.83 & 94.44 & 69.39 & 80.00 & 72.34 & 69.39 & 70.83 & 81.40 & 71.43 & 76.09 & 83.33 & 71.43 & 76.92 & 85.11 & 81.63 & 83.33 & 86.96 & 81.63 & \textbf{84.21}  \\  
\texttt{DETERMINATION}  & 100.00 & 36.36 & 53.33 & 85.71 & 54.55 & \textbf{66.67} & 100.00 & 36.36 & 53.33 & 83.33 & 45.45 & 58.82 & 85.71 & 54.55 & 66.67 & 83.33 & 45.45 & 58.82 & 42.86 & 27.27 & 33.33 \\
\texttt{DOC\_EVIDENCE}   & 77.61 & 74.29 & 75.91 & 77.27 & 72.86 & 75.00 & 80.60 & 77.14 & \textbf{78.83} & 75.00 & 72.86 & 73.91 & 68.42 & 74.29 & 71.23 & 67.53 & 74.29 & 70.75 & 71.62 & 75.71 & 73.61  \\  
\texttt{EXPLANATION}    & 46.00 & 43.40 & 44.66 & 60.98 & 47.17 & 53.19 & 56.82 & 47.17 & 51.55 & 58.14 & 47.17 & 52.08 & 53.49 & 43.40 & 47.92 & 54.17 & 49.06 & 51.49 & 60.47 & 49.06 & \textbf{54.17}  \\  
\texttt{GPE}            & 88.76 & 89.77 & 89.27 & 90.59 & 87.50 & 89.02 & 91.57 & 86.36 & 88.89 & 90.48 & 86.36 & 88.37 & 89.29 & 85.23 & 87.21 & 95.35 & 93.18 & \textbf{94.25} & 89.66 & 88.64 & 89.14  \\  
\texttt{LAW}            & 55.00 & 53.66 & 54.32 & 57.89 & 52.38 & 55.00 & 64.71 & 52.38 & \textbf{57.89} & 59.46 & 53.66 & 56.41 & 57.89 & 53.66 & 55.70 & 55.00 & 52.38 & 53.66 & 47.62 & 47.62 & 47.62  \\  
\texttt{LAW\_CASE}       & 71.43 & 33.33 & 45.45 & 66.67 & 26.67 & 38.10 & 71.43 & 33.33 & 45.45 & 46.15 & 40.00 & 42.86 & 50.00 & 46.67 & 48.28 & 56.25 & 60.00 & \textbf{58.06} & 37.50 & 40.00 & 38.71  \\  
\texttt{LAW\_REPORT}     & 100.00 & 66.67 & 80.00 & 66.67 & 66.67 & 66.67 & 100.00 & 66.67 & 80.00 & 66.67 & 66.67 & 66.67 & 100.00 & 66.67 & 80.00 & 50.00 & 66.67 & 57.14 & 66.67 & 66.67 & 66.67 \\
\texttt{NORP}           & 78.57 & 64.71 & 70.97 & 93.33 & 82.35 & 87.50 & 100.00 & 70.59 & 82.76 & 100.00 & 70.59 & 82.76 & 92.86 & 76.47 & 83.87 & 100.00 & 82.35 & 90.32 & 93.75 & 88.24 & \textbf{90.91} \\
\texttt{ORG}            & 64.71 & 67.35 & 66.00 & 78.38 & 59.18 & 67.44 & 73.81 & 63.27 & 68.13 & 73.33 & 67.35 & 70.21 & 78.57 & 67.35 & 72.53 & 80.39 & 83.67 & \textbf{82.00} & 82.93 & 69.39 & 75.56  \\  
\texttt{PERSON}         & 62.50 & 41.67 & 50.00 & 77.78 & 58.33 & 66.67 & 75.00 & 50.00 & 60.00 & 77.78 & 58.33 & 66.67 & 88.89 & 66.67 & 76.19 & 100.00 & 75.00 & \textbf{85.71} & 90.00 & 75.00 & 81.82 \\
\texttt{PROCEDURE}      & 71.67 & 69.35 & 70.49 & 73.77 & 72.58 & 73.17 & 71.93 & 66.13 & 68.91 & 73.77 & 72.58 & 73.17 & 76.67 & 74.19 & \textbf{75.41} & 71.01 & 79.03 & 74.81 & 74.58 & 70.97 & 72.73  \\

\bottomrule

\end{tabular}}
\caption{Precision (P), Recall (R) and F1 score (in \%) on the \textit{cover page} and the \textit{main text} for seven network architectures: baseline CNN model (\texttt{baseline}), CNN model with random static vectors on \texttt{en\_core\_web\_lg} (\texttt{CNN+rsv}), CNN with fine-tuned static vectors (\texttt{CNN+fts}), CNN with random static vectors and pretraining (\texttt{CNN+rsv+pt}), CNN with fine-tuned static vectors and pretraining (\texttt{CNN+fts+pt}), \texttt{RoBERTa}-based transformer, \texttt{LegalBERT}-based transformer}
\label{tab:results table}
\end{table*}

\begin{figure*}
   \includegraphics[width=\textwidth]{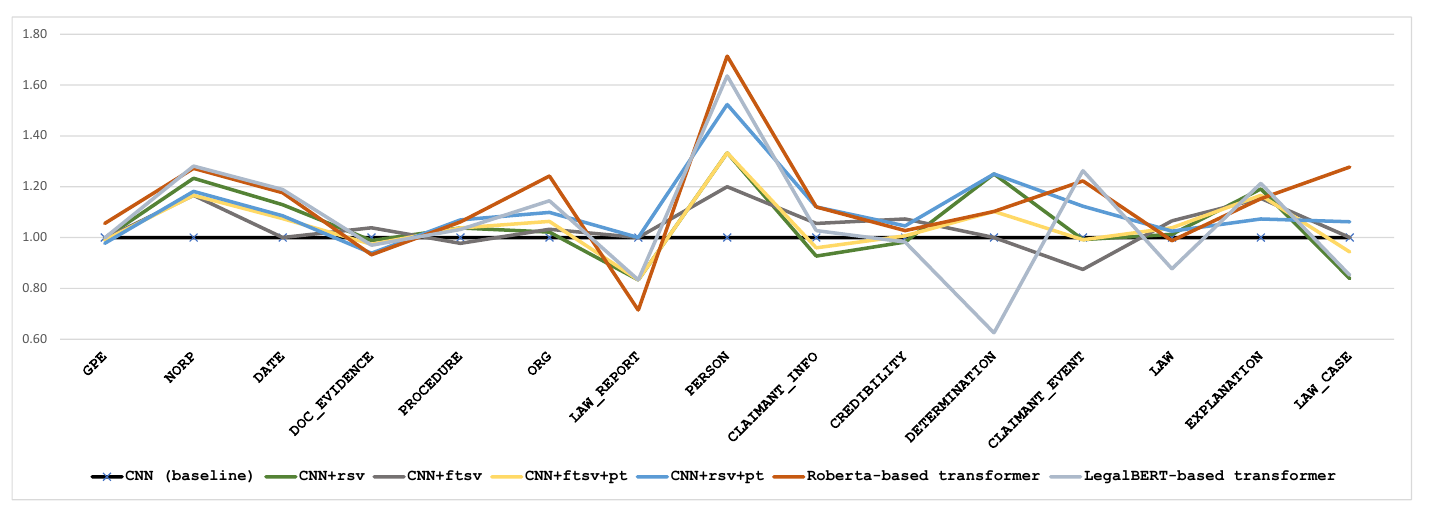} 
   \centering
    \caption{Comparison to the baseline on the F1 score on the \textit{main text}: per targeted category (x-axis) on seven network architectures: baseline CNN model (\texttt{baseline}), CNN model with random static vectors on \texttt{en\_core\_web\_lg} (\texttt{CNN+rsv}), CNN with fine-tuned static vectors (\texttt{CNN+fts}), CNN with random static vectors and pre-training (\texttt{CNN+rsv+pt}), CNN with fine-tuned static vectors and pre-training (\texttt{CNN+fts+pt}), \texttt{RoBERTa}-based transformer, \texttt{LegalBERT}-based transformer}
    \label{fig:baseline_graph}
\end{figure*}

\begin{figure}
   \includegraphics[width=\columnwidth]{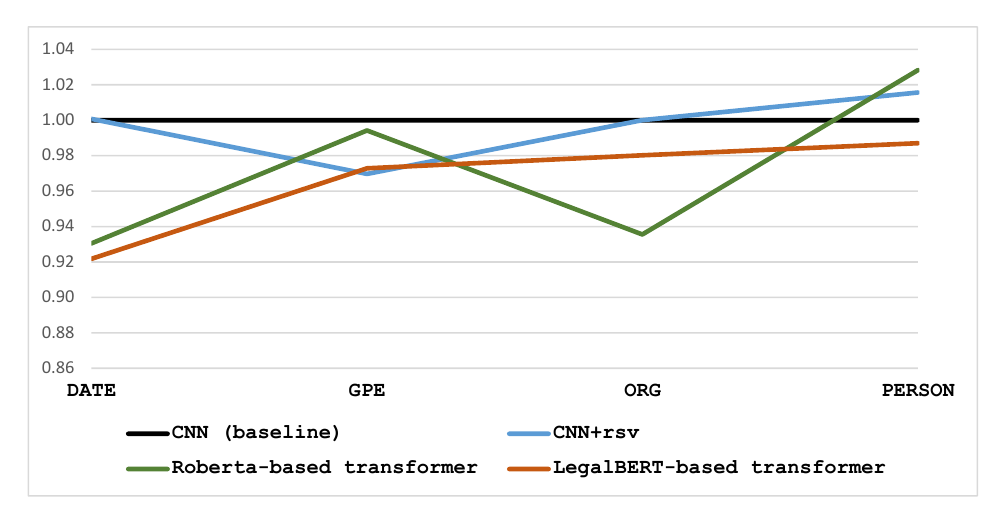} 
   \centering
    \caption{Comparison to the baseline on the F1 score on the \textit{case cover}: per targeted category (x-axis) on four network architectures: baseline CNN model (\texttt{baseline}), CNN model with random static vectors on \texttt{en\_core\_web\_lg} (\texttt{CNN+rsv}), \texttt{RoBERTa}-based transformer, \texttt{LegalBERT}-based transformer}
    \label{fig:baseline_graph_casecover}
\end{figure}

For the case cover, we obtain satisfactory results on all labels with F1 scores above 90\% for three of them and 84.78\% for name extraction. Apart from names, CNN architectures perform better, with dates achieving the highest score with randomly initialized embeddings. We explain this with the specific layout of this page (Annex \ref{sec:appendix A case cover}). The only gain of using a transformer-based model is to achieve a higher recall compared to the CNN-based architectures.

For the main text, results vary across labels: we obtain a score above 80\% for \texttt{DATE, GPE, PERSON, ORG} with the best score on \texttt{roBERTa}, but scores lower than 60\% on \texttt{EXPLANATION, LAW, LAW\_CASE}. Overall, when using transformers, we observe a better precision-recall balance. 

Results on three labels \texttt{DETERMINATION, LAW\_REPORT, NORP} are unreliable because of the limited sample both for training and testing. \texttt{DETERMINATION} appears only once per case, and \texttt{LAW\_REPORT} appears in a few cases only. Further annotation would require selecting the paragraphs of cases where these items appear to augment the size of the sample. We leave this task to future work. 

Explanations for other low scores are partly to be found in the tokenization errors reported during the human-labeling task. Figure \ref{figure: tokenization} shows an example of wrong tokenization on two categories  \texttt{LAW} and \texttt{LAW\_CASE} for which we believe bad tokenization is the primary explanation for low scores (similarly reported by \citeauthor{sanchez-2019-sentence}). In the first sentence of the figure, words are not correctly split between ``under'' and ``section'' and between the section number and ``of''. On the lower part of the figure, sentence tokenization does not correctly split the case reference as it is confused by the dot present in the middle. In this example, the case name is displayed as three different sentences, making the labeling task impossible.

The most appropriate pre-training varies across labels: For categories on which CNN performs best such as \texttt{CREDIBILITY, DOC\_EVIDENCE, LAW}, we find that fine-tuning static vectors performs better than randomly initialized embeddings or dynamic vectors, which suggests that context was not essential when retrieving spans of text (pre-training relies on tri-grams). This could derive from the methods of annotation that were terminology-based for those labels. While the target items may contain particular vocabulary such as ``personal information form'' for \texttt{DOC\_EVIDENCE}, context is of minimal importance since those phrases would not appear in another context or under another label. On the contrary, context seems much more important for retrieving procedural steps (\texttt{PROCEDURE}), which is the only category where the pre-training layer with contextual embeddings significantly increases the F1 score.

In the majority of categories, we find that the content of the pre-training is important (\texttt{CLAIMANT\_EVENT, CREDIBILITY, DATE, DOC\_EVIDENCE, EXPLANATION, LAW, PROCEDURE}). Results show that domain-specific training data has a larger effect than network architecture difference. More precisely, it seems that, on some categories (\texttt{CREDIBILITY, DOC\_EVIDENCE, LAW, PROCEDURE}), pre-training on our own data is more effective than training on a general legal data set as in \texttt{LegalBERT}. This can be explained by the content \texttt{LegalBERT} is pre-trained on, which does not contain any Canadian but only US, European, and UK texts and does not include any refugee cases. 

In other categories, \texttt{roBERTa} performs better than \texttt{LegalBERT} and CNNs, suggesting that the size of the pre-trained model is more important than domain matching. While \texttt{LegalBERT} has a size of 12GB, \texttt{roBERTa} is over 160GB and outperforms \texttt{LegalBERT} on traditional NER labels (\texttt{GPE, ORG, PERSON} and also \texttt{CLAIMANT\_INFO, LAW\_CASE}). 

Looking at recall measures only, the superiority of transformer architectures against CNNs is more significant, with only 3 categories (\texttt{DOC\_EVIDENCE, CLAIMANT\_INFO, LAW}) achieving their best recall score with a CNN architecture and legal pre-training. Comparing results on recall, we reach the same conclusion as with F1, i.e.\ that domain matching allows us to achieve higher scores on target categories. Indeed, for seven out of 12 categories analyzed for the main text, the best scores are achieved by two architectures that differ in their pre-training domain. Higher F1 and recall scores, obtained through comparison and observation, enable us to attribute the improved performance primarily to the domain of the training data.

\section{Related work}
Because of the importance of language and written text, applications of NLP in law hold great promise in supporting legal work, which has been extensively reviewed by \citeauthor{zhong-etal-2020-nlp}. However, because of the specificity of legal language and the diversity of legal domains, as demonstrated in our work with the results on \texttt{LegalBERT}-based transformer, general approaches aiming at structuring legal text such as \textit{LexNLP} \cite{bommarito2021lexnlp} or general legal information extraction \cite{bruninghaus2001improving} are unfit for specific domains such as international refugee law and are not able to achieve a high degree of granularity. 

Earlier methods of statistical information extraction in law include the use of linear models such as maximum entropy models \cite{bender2003maximum, clark2003combining} and hidden Markov models \cite{mayfield2003named}. However, state-of-the-art results are produced by methods able to capture some context, with an active research community investigating the use of conditional random fields \cite{benikova2015c, faruqui2010training, finkel2005incorporating} and BiLSTMs \cite{chiu2016named, huang2015bidirectional, lample-etal-2016-neural, ma-hovy-2016-end, Leitner_Elena} for legal applications.

Scope and performance increased with the introduction of new architectures of deep learning using recurrent neural networks (RNN), CNNs, and attention mechanisms as demonstrated by \citeauthor{chalkidis-etal-2019-extreme}, even though we find that transformers do not always perform best on our data. We therefore focus in this work on statistical NER approaches. Attempts have been made to extract legal elements of the procedure using \texttt{spaCy} CNNs \cite{pais-etal-2021-named, Vardhan_NER} with the latter achieving a total F1 score of 59.31\% across labels, citations, as well as events, which is below our reported scores. 

Similar case matching is a well-known application of NLP methods, especially in common law systems \cite{trappey_identify_2020} and in domains such as international law. The Competition on Legal Information Extraction/Entailment includes a task of case retrieval, which proves that there is much interest in this area both from researchers and the developers of commercial applications. While research has been conducted to match cases at paragraph level \cite{tang-clematide-2021-searching, hu2022bert_lf}, we find that our approach is more transparent and shifts the decisions regarding which filters to choose to legal practitioners, which we believe is appropriate to enable productive human-machine collaboration in this high-stakes application domain.


\section{Conclusion and future work} 

Our pipeline identifies and extracts diverse text spans, which may vary in quality across different categories. We acknowledge that certain entities we identify are more valuable than others for legal search purposes. Additionally, due to the complexity of the text, some noise is to be expected. However, this noise does not hinder the search for relevant items. Users have the flexibility to search and retrieve cases using any combination of our 19 categories of flagged entities. Additionally, work is required for the evaluation of the prototype by legal practitioners beyond traditional machine learning metrics \cite{barale2022human}. However, we believe the work presented here is an important first step and has the potential to be used for future NLP applications in refugee law. 
Our approach provides significant contributions with newly collected data, newly created labels for NER, and a structure given to each case based on lawyers' requirements, with nine categories of information being retrieved with an F1 score higher than 80\%. Compared to existing case retrieval tools, our pipeline enables end-users to decide what to search for based on defined categories and to answer the question: \textit{What are criteria of similarity to my new input case ?}



\section*{Limitations}
In this section, we enumerate a few limitations of our work: 
\begin{itemize}
    \item We believe that the need to train transformer architectures on GPU is an obstacle to the use of this pipeline, which is destined not to be used in an academic environment but by legal practitioners. 
    \item Because of the specificity of each jurisdiction, generalizing to other countries may not be possible on all labels with the exact same models (for example in extracting the names of tribunals).
    \item The manual annotation process is a weakness: while it results in gold-standard annotations, it is very time-consuming. We do acknowledge that the amount of training data presented in this work is low and that collecting more annotations in the future would improve the quality of the results. We think it would be interesting to look at self-supervised methods, weak supervision, and annotation generation. The need for labeled data also prevents easy replication of the pipeline to new data sets, which would also require manually annotating. 
    \item More precisely on the extracted categories, some categories lack precision and would require additional processing steps to achieve satisfactory results. For example, the category \texttt{PERSON} sometimes refers to the claimant or their family, but sometimes refers to the name of the judge. 
\end{itemize}

\bibliography{custom}
\bibliographystyle{acl_natbib}

\appendix

\section{Example of a \textit{case cover}}
\label{sec:appendix A case cover}

\begin{figure*}
   \includegraphics[width=\textwidth]{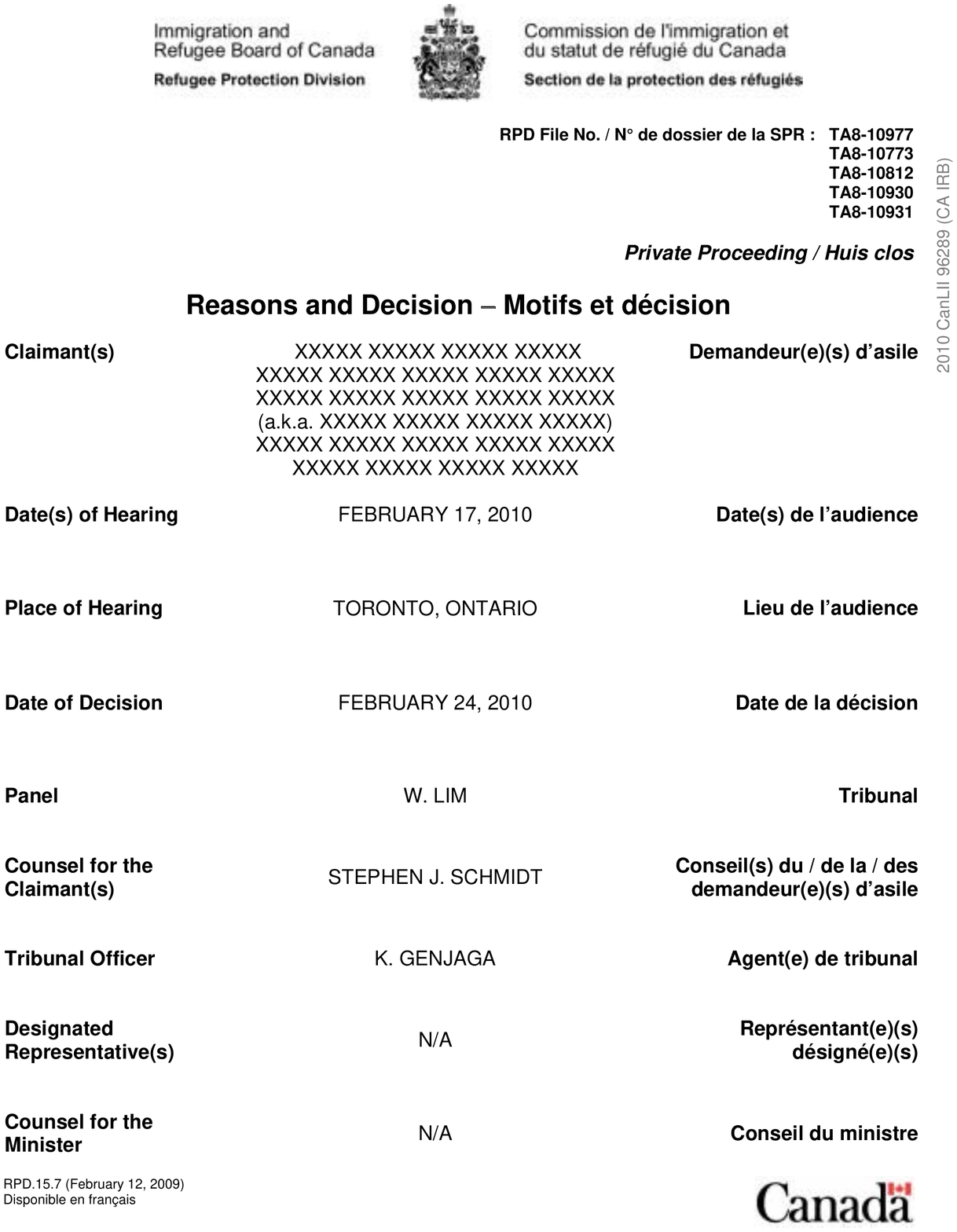} 
   \centering
   \caption{Example of a \textit{case cover}}
   \label{figure: case cover}
\end{figure*}


\end{document}